\documentclass[letterpaper]{article}
\usepackage{amsmath,epsfig}
\usepackage[preprint]{spconfa4}

\let\OLDthebibliography\thebibliography
\renewcommand\thebibliography[1]{
  \OLDthebibliography{#1}
  \setlength{\parskip}{0pt}
  \setlength{\itemsep}{0pt plus 0.3ex}
}

\usepackage{algorithm}
\usepackage{algorithmic}

\usepackage{booktabs}
\usepackage{amssymb,amsfonts}

\begin{document}\sloppy
\topmargin=0mm
\def\x{{\mathbf x}}
\def\L{{\cal L}}

\title{Towards Better Graph Representation: Two-Branch Collaborative Graph Neural Networks for Multimodal Marketing Intention Detection}

\name{Lu Zhang, Jian Zhang, Zhibin Li, and Jingsong Xu}
\address{University of Technology Sydney, Australia\\ \{lu.zhang-5@student., jian.zhang@, zhibin.li@student., jingsong.xu@\}uts.edu.au}

\maketitle

\begin{abstract}
Inspired by the fact that spreading and collecting information through the Internet becomes the norm, more and more people choose to post for-profit contents (images and texts) in social networks. Due to the difficulty of network censors, malicious marketing may be capable of harming the society. Therefore, it is meaningful to detect marketing intentions online automatically. However, gaps between multimodal data make it difficult to fuse images and texts for content marketing detection. To this end, this paper proposes Two-Branch Collaborative Graph Neural Networks to collaboratively represent multimodal data by Graph Convolution Networks (GCNs) in an end-to-end fashion. We first separately embed groups of images and texts by GCNs layers from two views and further adopt the proposed multimodal fusion strategy to learn the graph representation collaboratively. Experimental results demonstrate that our proposed method achieves superior graph classification performance for marketing intention detection. 
\end{abstract}


\section{Introduction}
Nowadays, the Internet has become one of the most powerful mediums for marketing dissemination. Social media, such as Facebook, Twitter, and Sina Weibo, acts as an open platform on which a massive amount of self-media accounts or persons post texts with images every day. With the explosion of media data online, plenty of contents with marketing intentions appear. To release the disgust of readers and largely gain trusts of potential customers, marketing contents usually conceal in normal articles for obtaining benefits. Due to the absence of network surveillance, a lot of malicious marketing information, such as fake news, misleading advertisements, and unproven folk science, disseminates among the crowds. These malicious marketing contents may be capable of harming social order, people's economic interest, and even health. Even normal promotional advertisement can distract people's attention and waste their time. Facing this phenomenon, automatically marketing intention detection is of great importance~\cite{bhosale2013detecting}. 

Textual marketing contents are in the format of texts when spreading. Some prior works~\cite{article,wang2019stacking} use several types of textual features, and deep learning embedding features to mine marketing intentions of news. These approaches follow the Natural Language Processing (NLP) algorithms to solve the problem of marketing intention detection. They focus on mining the semantic information in a single modality and ignore the combination and guidance of multiple modalities. Nowadays, marketing contents are in the format of a mixture of texts and images and even come with groups of both. 
A variety of applications already appear to collaboratively use multimodal data to improve their task performance, such as sentiment analysis~\cite{zhang2019modeling}, and image captioning~\cite{meng2019multimodal}, since data from different modality lead to a complement of each other. 
Some works~\cite{harzig2018multimodal,harzig2019image} concentrate on automatically generating descriptive captions for images for the purpose of marketing.
However, to our best knowledge, there is limited work focusing on multimodal marketing content detection, especially when a piece of news contains multiple images. 

In this study, we focus on semantic-based multimodal marketing intention detection involving texts and images. We embed multiple modalities by Graph Convolution Networks (GCNs) and jointly extract semantic information to recognize marketing intention contents. 
One of the biggest challenges in the task of content marketing detection is how to dig deep latent semantic information of multiple modalities and fuse them via embedding. 
Especially when marketing contents are embedded in normal articles, it becomes difficult to identify marketing intentions among the interference of other normal topics. 
Moreover, a piece of news usually includes variable numbers of images and sentences at the same time. This brings difficulties for neural networks to process and integrate data. 

\begin{figure*}[htp]
	\begin{center}
		\includegraphics[scale=0.55]{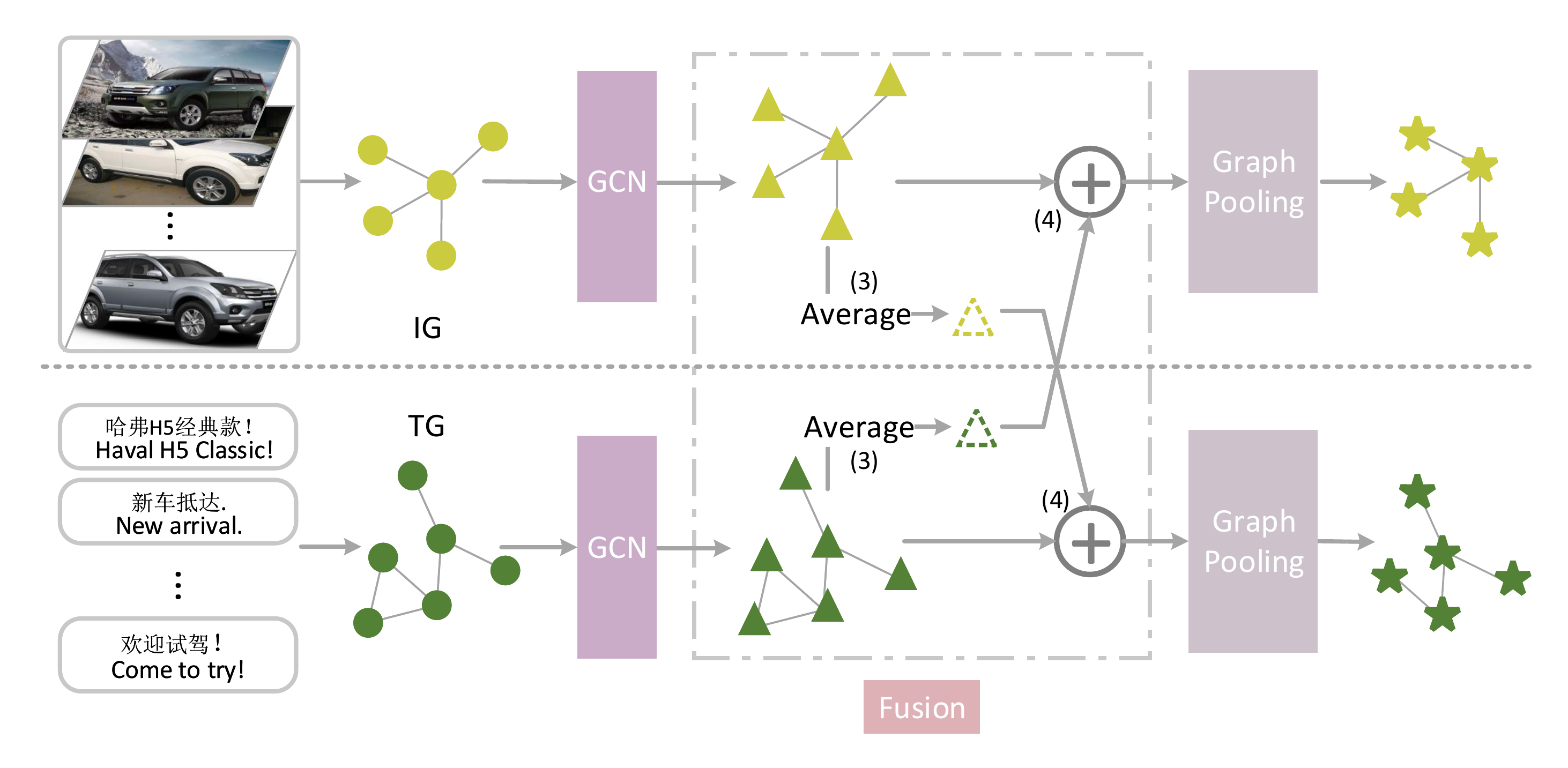}
		\caption{The structure of our proposed multimdal fusion strategy. Images and texts are represented as graph structures and then are fed into two GCNs. Then we calculate the graph center and crosswise accumulate. Graph pooling layer is followed after fusion.}\label{bigFig}
	\end{center}
\end{figure*}

To overcome the challenges above, this paper proposes to learn an image-text collaborate embedding by using a two-branch GCNs framework, as shown in Figure~\ref{bigFig}. We first construct two graph structures by integrating groups of images/texts: the image graph and the text graph. The motivation is that images/texts of the same piece of news have strong or weak intra modality correlations with each other. Graph structure is good at representing this kind of relations. 
The image graph is composed of images in the same piece of news, and the text graph is composed of sentences. Edges are built by the similarity between nodes. The number of nodes in each graph varies a lot since the numbers of images/sentences in pieces of news are different. Because of the heterogeneity of multimodal, we then use two-branch GCNs to mine inter relationships between images and texts. 

Our fusion strategy is as follows: We extract the embedded features of the graph convolution layer and calculate the average as the graph center of each modality, as shown in Figure~\ref{bigFig}. We then add the graph center of image/text graph to each node in text/image graph. The accumulation results will be fed into the next graph convolution layer. We use the extracted image/text graph center as a guidance of the other modality for the next convolution. Multimodal information exchanges after every convolution layer.
By repeatedly doing this, similar semantic information will be enhanced, and disparate semantic information will be weakened. We also leverage graph pooling layers to refine the graph representation further. Finally our framework jointly learns the embedded features of cross-modal data containing evidence as much as possible for marketing intention identify. We conduct a variety of experiments on a real dataset. The results demonstrate our superior performance on this task compared with state-of-the-art methods. 

In general, our main contributions can be summarized as follows:
\begin{itemize}
  \item We propose to use the graph structure to represent flexible numbers of images and texts of news and extract the combined multimodal features to detect the marketing intentions of social media automatically.
  \item We propose a multimodal collaborative fusion strategy after every convolution layer in two-branch GNNs in an end-to-end fashion. We achieve a better integration of multimodal data. 
  \item We conduct a number of experiments. The results demonstrate the superior performance of the proposed method.
\end{itemize}

\section{Related Work}
In this section, we briefly review the related works in these aspects: traditional marketing detection, multimodal embedding and graph representation. 

\textbf{Traditional marketing detection.} 
In the literature, early studies follow the traditional NLP view to define the content marketing detection as a simple textual classification task and incorporate traditional low-level textual features extracted by TF-IDF, Word2Vec and so on.~\cite{bhosale2013detecting} uses stylometric features based on both n-grams and Probabilistic Context Free Grammars (PCFGs) which improves over using only shallow lexical and meta-features.

\textbf{Multimodal embedding.} 
As one of the representative statistic methods, Canonical Correlation Analysis (CCA)~\cite{gao2019labeled} embeds multimodal data by exploring the relationship between two sets of variables. CCA maximizes the data correlation and achieves the linear projection. While in reality, linear projection does not gain enough capability to represent all attributes. Considering this aspect, deep learning methods perform better in an end-to-end fashion. Some works~\cite{peng2018ccl} learn deep embedding by metric loss function to make sure the distance of multimodal data with similar semantic information will be short. ~\cite{wang2018learning} uses maximum-margin ranking loss and novel neighborhood constraints to learn two-branch Convolution Neural Networks (CNNs). The example sampling strategy leads to high computation complexity. 

\textbf{Graph representation.} 
CNNs or Recurrent Neural Networks (RNNs) successfully leverage the properties of data such as images, texts and videos on Euclidean domain~\cite{lee2019self}. While in real world, a variety of non-Euclidean data exist, such as social network and molecular structures. GNNs is a type of neural networks which is capable to directly operated on such graph structure.
GNNs has two branches: the spectral methods and the non-spectral methods. GCNs~\cite{kipf2016semi,zhou2018graph} is one of the representative spectral methods which has already shown its outstanding performance in a variety of applications. GCNs focuses on local connection among graph nodes in a shared weights strategy by a multi-layer structure. Non-spectral methods involve aggregation and combination~\cite{xu2018powerful}, such as GraphSAGE~\cite{hamilton2017inductive} and GIN~\cite{xu2018powerful}. Based on GNNs, a lot of graph attention~\cite{velivckovic2017graph} and graph pooling~\cite{ying2018hierarchical,lee2019self} methods are proposed. 

\section{Proposed Method}
  
In this section, we define our problem of multimodal content marketing detection and describe our Two-Branch Collaborative GNNs. 

\subsection{Problem Definition}
Assume a piece of news composed of $n$ images  $<x_1^{(I)},x_2^{(I)},...,x_n^{(I)}>$  and $m$ description sentences $<x_1^{(S)},x_2^{(S)},...,x_m^{(S)}>$ with superscript $I$ and $T$ standing for image and sentence respectively. We generate two graph structures including ImageGraph $IG$ and TextGraph $TG$ as shown in Figure~\ref{bigFig}. Each graph consists of two components: nodes and edges. In ImageGraph, each image is a node and edges are constructed by similarities between each pair of images. The definition is similar to TextGraph. We denote $IG = (x^{(I)}, b^{(I)})$ and $TG = (x^{(S)},b^{(S)})$, in which $b^{(I)}$ refers to edges in $IG$ and $b^{(S)}$ refers to edges in $TG$. We define $b_{ij}$ as the undirected edge between nodes $i$ and $j$.
We use features generated by pre-trained Resnet18~\cite{he2016deep} to build our edges in ImageGraph. We calculated the similarity between nodes:
\begin{eqnarray}
S_{ij} = \frac{x_i x_j}{\|x_i\| \|x_j\|}.
\end{eqnarray}
\begin{figure}[!htp]
	\begin{center}
		\includegraphics[scale=0.37]{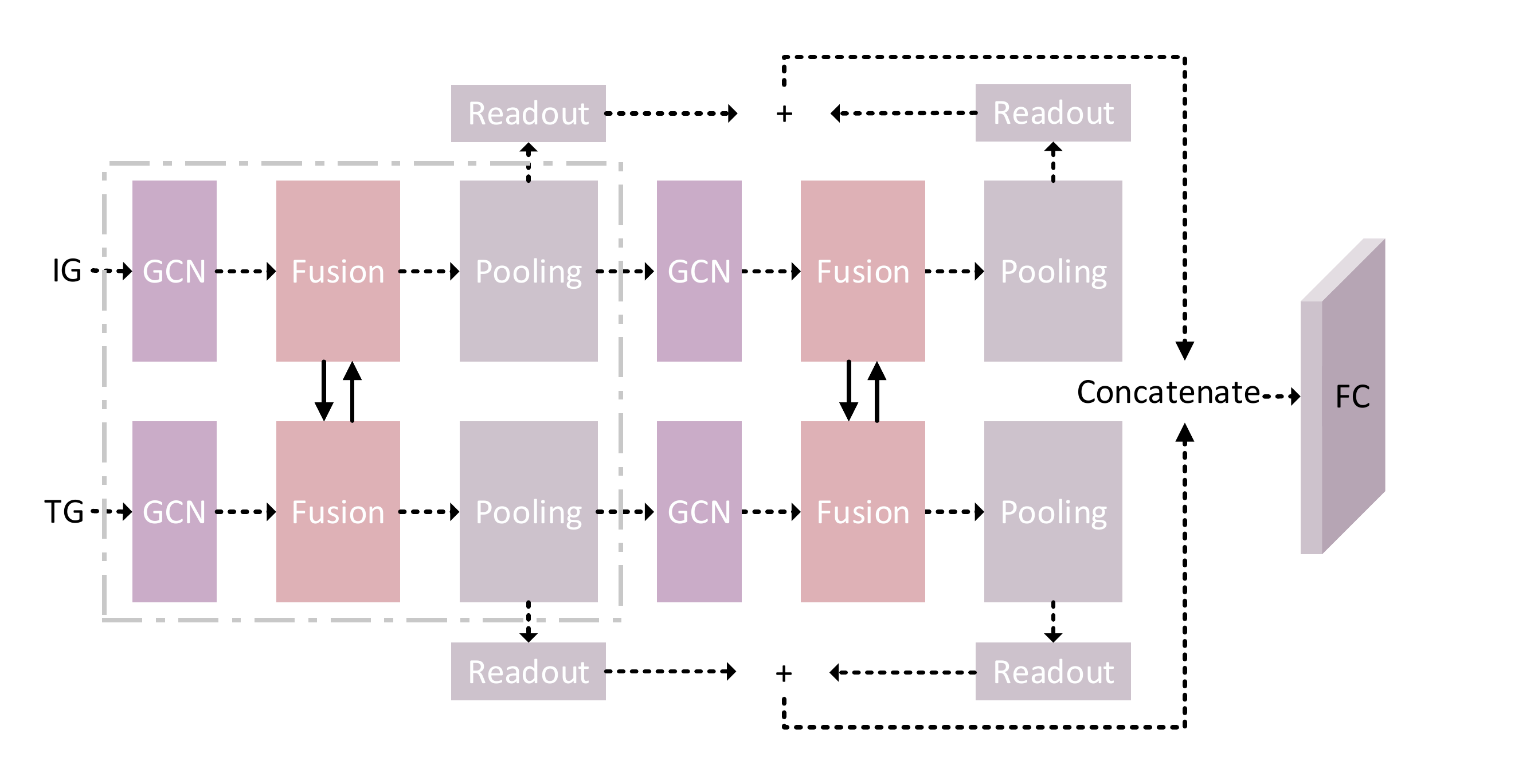}
		\caption{Network architecture.}\label{smallFig}
	\end{center}
\end{figure}

We put edge weight one if the cosine similarity between two nodes is in the top 50\% among all cosine similarities. A similar idea is applied for the construction of TextGraph with Word2Vector features. 

\subsection{Two-branch Collaborative Graph Neural Networks}
\textbf{Methodology.} Because of the heterogeneity of different modality data, some works~\cite{wang2018learning} already proposed to use two-branch CNNs to solve multimodal jointly classification or retrieval problems. However, the two branches are lack of interaction but just concatenated or summarized at the end layer. These methods separately learn to embed images and texts without supplementary guidance or fusion. To ensure good performance, they have to constrain representation results by metric loss, which may need strong computation capability. On the premise of the absence of metric loss, our Two-branch Collaborative GNNs use the embedding results of the previous convolution layers to guide the next convolution. 
We introduce the image branch here for example. The branch for text can be constructed similarly.
Concretely, we calculate the image graph center of the last layer and use it to update the text nodes iteratively. 
Through this, similar semantic information will be enhanced. At the meantime semantically disparate information will be weakened.
For further enhancing the representation capability, we employ graph pooling layers. 
Then we use the updated text nodes to conduct the next convolution. 

\begin{table*}[t]
\begin{center}
\caption{We construct ImageGraph and TextGraph and random split into training, validation and testing sets.} \label{tab:dataset}
\begin{tabular}{@{}ccccccc@{}}
\toprule
               & \# ImageGraph & \# Image nodes & \# Image edges & \# TextGraph & \#Text nodes & \#Text edges \\ \midrule
Totally        & 12375 & 58228  & 69215 & 12375 & 319504 & 5291650 \\
Training set   & 9900  & 46515  & 55231 & 9900  & 252711 & 3568798\\  
Validation set & 1238  & 5885   & 7027  & 1238  & 32033  & 482728    \\  
Testing set    & 1237  & 5828   &6957   & 1237  &  34760 & 1240124  \\ 
\bottomrule
\end{tabular}
\end{center}
\end{table*}

The layer-wise propagation rule~\cite{kipf2016semi} is:
\begin{eqnarray}
H_{l+1} = \sigma(\widetilde{D}^{-\frac{1}{2}} \widetilde{A} \widetilde{D}^{-\frac{1}{2}} H_{l} \Theta_{l}).
\end{eqnarray}
where $H_{l} \in \mathbb{R}^{N \times C}$ is the matrix of activations in the ${l}^{th}$ convolution layer with $N$ nodes. 
$C$ is the feature dimension.
$\sigma$ is the activation function such as ReLU($\cdot$)=max(0,$\cdot$). 
$\widetilde{A} = A + I_N$, where $A$ is the adjacency matrix, $I_N$ is the identity matrix.
$\widetilde{D}$ is the degree matrix of $\widetilde{A}$: $\widetilde{D}_{ii} = \sum_j{\widetilde{A}_{ij}}$.   
$\Theta_{l}$ is the only trainable weighted matrix. In the first layer, $H_{0}=X$. 

Suppose $H^{(I)}_{l}$ and $H^{(S)}_{l}$ are outputs of the $l^{th}$ layers of image branch and text branch. Image graph centre $ \overline{h}^{(I)}_{l} \in \mathbb{R}^{1 \times C}$ is calculated as:
\begin{eqnarray}
\overline{h}_l^{(I)} = \frac{ \mathbf{1}_{1 \times {N_1}} H_l^{(I)}}{N_1}.
\end{eqnarray}
Here $\mathbf{1}_{1 \times {N_1}}$ is a vector of all ones. $N_1$ is the number of image nodes after the $l^{th}$ layer.
The update formula of text nodes is:
\begin{eqnarray}
{H_l^{(S)}}' = H_l^{(S)} + \mu \overline{H}_l^{(I)}.
\end{eqnarray}
Here $\mu$ is a learnable parameter to adjust the fusion weight. $\overline{H}_l^{(I)} \in \mathbb{R}^{{N_2} \times C}$ in which every row is $\overline{h}_l^{(I)}$.
We add a graph pooling layer $f_p$ after the accumulation.
For the next convolution layer:
\begin{eqnarray}
H_{l+1}^{(S)} &= &\sigma(\widetilde{D}^{-\frac{1}{2}} \widetilde{A} \widetilde{D}^{-\frac{1}{2}} f_p({{H_l^{(S)}}'}) \Theta_{l})   \nonumber \\
&= &\sigma(\widetilde{D}^{-\frac{1}{2}} \widetilde{A} \widetilde{D}^{-\frac{1}{2}} f_p{(H_l^{(S)}+\mu \overline{H}_l^{(I)})} \Theta_{l}).
\end{eqnarray}

\textbf{Network architecture.} We define the networks in Figure~\ref{bigFig} as a basic block. Each branch in our framework passes the data through two basic blocks, followed by a fully connection, as shown in Figure~\ref{smallFig}. Each block contains a GCN layer, a data fusion layer and a pooling layer. The fusion layer is our proposed node update layer. The multimodal information exchanges in this layer. We implement the Self-Attention Graph Pooling from~\cite{lee2019self}. In Readout layer, we aggregate all nodes of the graph to represent the whole graph. In detail, we calculate the max and average of all nodes and then concatenate them together:
\begin{eqnarray}
R_l = Max(f_p({H_{l}}')) \| Average(f_p({H_{l}}')).
\end{eqnarray}
Here $\|$ means concatenate. The last layer is a fully connection layer for graph classification.  

\section{Experiments}
In this section, we evaluate our proposed Two-Branch Collaborative GCNs with other baselines. We describe the dataset we used and our experiment settings in detail. 

\subsection{Dataset Description}

\begin{table}[]
\begin{center}
\caption{Comparison with five baselines.} \label{tab:baselines}
\begin{tabular}{@{}c|c|c@{}}
\hline
Method       & Loss  & Accuracy \\
\hline
\hline
Image-branch GNN  & 0.620  & 0.645         \\
 Text-branch GNN   & 0.584 & 0.689    \\
Two-branch SN~\cite{wang2018learning}      & 0.568 & 0.698    \\
Two-branch GNN     & 0.565 & 0.707    \\
Two-branch GNN + SAGPool~\cite{lee2019self} & 0.558 & 0.712   \\
\textbf{Ours}    & 0.534  & \textbf{0.740}   \\
\hline
\end{tabular}
\end{center}
\end{table}

\textbf{Data construction.} We use dataset released from SOHU's second content recognition algorithm competition~\cite{Web}. The dataset includes 50000 tagged news. We filter news containing images less than 3 and sentences less than 5, because we focus on news with multiple images and relatively long text.  Considering the computational limitation, we also abandon news with more than 8 images. 
Totally we get 12375 news and build 12375 ImageGraphs and 12375 TextGraphs. The number of samples with marketing intentions is 7258, and the number of samples without marketing intention is 5117. We construct 69215 undirected edges and 58228 image nodes in $IG$ and 5291650 undirected edges and 319504 text nodes in $TG$. We random split 12375 samples into 9900 training set, 1237 testing set, and 1238 validation set. Detailed information is shown in Table~\ref{tab:dataset}. 

\textbf{Data pre-process.} For the text data, we firstly delete some useless punctuations in our corpus. Since the dataset is a Chinese dataset, we use `jieba' Chinese word segmentation tool~\cite{jieba} and filer the stop words. We leverage the Word2vec~\cite{P18-2023} model to extract the raw textual features of dimension 256. For image feature extraction, we use the ResNet18~\cite{he2016deep} model with output dimension 256.  
\begin{figure*}[htp]
	\begin{center}
		\includegraphics[scale=0.6]{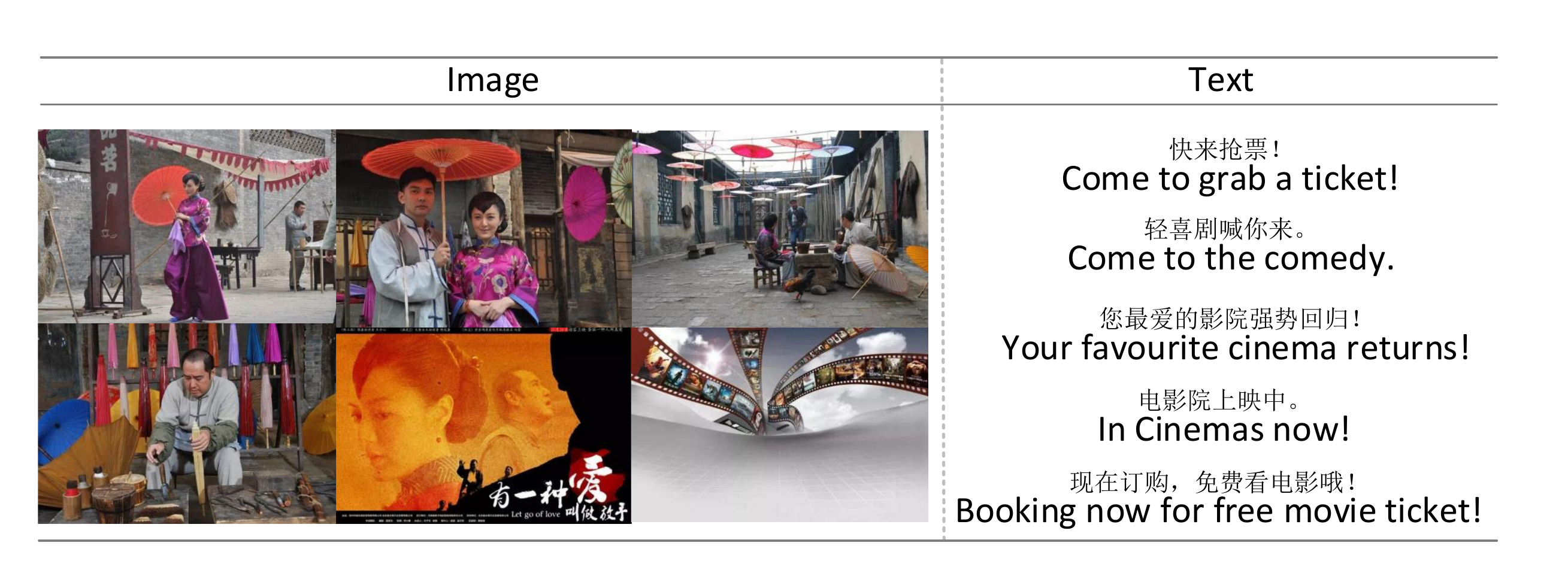}
		\caption{An example in the dataset. From the texts, we can easily identify the marketing intention. However from only images, the topic is blurry.}\label{show}
	\end{center}
\end{figure*}

\subsection{Experimental Setup}
Image branch and text branch use the same network configurations. The output feature dimensions of the first GCNs layer is 64. The second GCNs layer is 32. Pooling ratio is 0.8. The weight matrix of last three fully connection layer is $128 \times 64$, $64 \times 32$ and $32 \times 2$. We optimize our networks by the Adam~\cite{kingma2014adam} optimizer with an initial learning rate of 0.001. Batch size is 32. Weight decay is 0.000001.

We compare to five baselines: Text-branch GNN, Image-branch GNN, Two-branch SN, Two-branch GNN, and Two-branch GNN + SAGPool. Text-branch GNN and Image-branch GNN are single-branch methods that only use the text/image data for marketing intention detection. Two-branch SN, Two-branch GNN, and Two-branch GNN + SAGPool methods use multimodal data in two-branch neural network frameworks for the task. 

Text-branch GNN method uses a single text branch network without any image information for classification.
Image-branch GNN method uses a single image branch network without any text information for classification.
Two-branch SN is the two-branch Similarity Network proposed in~\cite{wang2018learning}. However, this method is designed for matching a image/text for an text/image, which cannot be applied directly to our multiple image and text data classification. We average the embedded features of all images/texts in a piece of news to represent the overall image/text feature. 
Two-branch GNN is the two-branch GNNs with on pooling layers for the representation of multiple images and texts in each piece of news.
Two-branch GNN + SAGPool is the two-branch GNNs with SAGPooling layers~\cite{lee2019self}.

\subsection{Experimental Results}

Table~\ref{tab:baselines} compares the logloss (cross entropy) of binary classification and accuracy scores of the proposed method with the other five baselines. The last row is our proposed method. 

The first two single-branch methods perform the worst, due to the lack of multimodal information. By involving multimodal data, the three two-branch baselines perform better. Our proposed method fuses multimodal data after every convolution layer, which achieves the best performance. 
Concretely, the performance of the Image-branch GNN method is inferior to other methods. The loss is also the highest. The accuracy score is 0.645. This is because images always supplement the vivid information in a piece of news. Words still play a leading role in information broadcasting. Sometimes it is not clear or even misleading for people by only seeing images. For example, if there is a piece of news in which the content is about movie propaganda, as shown in Figure~\ref{show}. Some images are just film stills. Without the textual information, it is hard for readers to recognize marketing intention. Because film subjects vary a lot, visual images lead people directly into the movie scenes. Text-branch GNN method performs better than Image-branch GNN method. The accuracy score improves 0.044. This bigger enhancement indicates the importance of texts. While due to the lack of multimodal fusion, the performance is worse than other baselines. Two-branch SN improves a little comparing to the previous two baselines. 
Two-branch GNN and Two-branch GNN + SAGPool achieve similar performance, whose accuracy scores are around 0.71. It is reasonable that these two approaches are both based on two-branch GNNs. Overall, methods involving multimodal data improve the performance compared with single-modal methods. 

From the results, it is easy to observe that our proposed method achieves significant improvements on the dataset in marketing intention detection. Our method performs not only better than the single-modal methods but also better than other multimodal methods. By adding fusion layers in the framework, similar semantic information will be enhanced, and disparate semantic information will be weakened.
The proposed method jointly learns the embedded features of cross-modal data containing evidence as much as possible to identify the marketing intentions.

\section{Conclusion}
In this paper, we propose Two-Branch Collaborative Graph Neural Networks for marketing intention detection of multimodal data. In our framework, considering the heterogeneity of texts and images, we add a fusion layer between two convolution layers for better graph representation. The experiment results demonstrate the effectiveness of the proposed method for marketing intention detection. 




\bibliographystyle{IEEEbib}
\bibliography{icme2020template}

\end{document}